\documentclass[12 pt, onecolumn]{IEEEtran}
\usepackage[utf8]{inputenc} 
\usepackage[T1]{fontenc}
\usepackage{amsfonts}
\usepackage[cmex10]{amsmath} 
\usepackage{amsthm}
\usepackage{url}
\usepackage{lettrine}
\usepackage{framed}
\usepackage{color}
\usepackage{algorithm}
\usepackage{algpseudocode}
\usepackage{dsfont}
\usepackage{booktabs} 
\usepackage{framed}
\usepackage{bm}
\usepackage{bbm}
\usepackage{ifthen}
\usepackage{enumitem}
\usepackage{cite}
\usepackage{graphicx}
\usepackage{epstopdf}
\usepackage[section]{placeins}
\usepackage{subcaption}
\usepackage{dsfont}
\usepackage{framed}
\usepackage{bm}
\usepackage{bbm}
\usepackage{helvet}
\usepackage{xcolor}
\newtheorem{theorem}{Theorem}

\newtheorem{lem}{Lemma}
\usepackage{xr}

\newcommand{\abs}[1]{\left|#1\right|}

\newcommand{\real}{\mathbb{R}}
\newcommand{\prob}[1]{\mathbb{P}\left(#1\right)}

\newcommand{\bvec}[1]{\bm{#1}}
\newcommand{\norm}[1]{\left\|#1\right\|_2}

\newcommand{\opnorm}[2]{\left\|#1\right\|_{#2}}

\newcommand{\proj}[1]{\bm{P}_{#1}}

\DeclareMathOperator*{\argmin}{\arg\,\min}

\IEEEoverridecommandlockouts
\title{Follow The Approximate Sparse Leader for No-Regret Online Sparse Linear Approximation}
\begin{document}
\author{%
  \IEEEauthorblockA{Samrat Mukhopadhyay, \emph{Member, IEEE}, and, Debasmita Mukherjee}
 \thanks{The authors are with the Dept. of Electronics Engineering, IIT (ISM) Dhanbad, Jharkhand, India. This research is supported by the Science and Engineering Research Board (SERB), India, Grant no: SRG/2022/001781}
}
\maketitle
\setlength{\abovedisplayskip}{0pt}
\setlength{\belowdisplayskip}{0pt}
\begin{abstract}
	\label{sec:abstract}
	We consider the problem of \textit{online sparse linear approximation}, where one predicts the best sparse approximation of a sequence of measurements in terms of linear combination of columns of a given measurement matrix. Such online prediction problems are ubiquitous, ranging from medical trials to web caching to resource allocation. The inherent difficulty of offline recovery also makes the online problem challenging. In this letter, we propose Follow-The-Approximate-Sparse-Leader, an efficient online meta-policy to address this online problem. Through a detailed theoretical analysis, we prove that under certain assumptions on the measurement sequence, the proposed policy enjoys a data-dependent sublinear upper bound on the static regret, which can range from logarithmic to square-root. Numerical simulations are performed to corroborate the theoretical findings and demonstrate the efficacy of the proposed online policy.  
\end{abstract}
\begin{IEEEkeywords}
	Sparse Approximation, Static Regret, Follow-The-Leader (FTL).
\end{IEEEkeywords}
\section{Introduction}
\label{sec:Introduction}
Sparse linear approximation is the problem of finding the (specific number of) columns of a given measurement matrix that best linearly approximates a given measurement vector. Mathematically, the canonical offline problem of sparse linear approximation is to find $\bm{x}\in \real^N$ such that $\bm{x}$ is $K-$sparse, i.e., $\bm{x}$ has at most $K$ non-zero entries, such that $\norm{\bm{y}-\bm{\Phi x}}$ is minimized for a given measurement vector $\bm{y}\in \real^M$ and a measurement/sensing matrix $\bm{\Phi}\in \real^{M\times N}$. In the past two decades, an array of specialized optimization approaches, ranging from convex relaxation techniques, such as LASSO~\cite{tibshirani1996regression}, Basis pursuit~\cite{chen2001atomic}, Dantzig selector~\cite{candes2007dantzig}, etc., and greedy sparse recovery algorithms, such as IHT~\cite{daubechies2004iterative}, SP~\cite{dai2009subspace}, CoSaMP~\cite{needell2009cosamp}, HTP~\cite{foucart2011hard}, etc., have been proposed, which enjoy provable exact recovery guarantees if the measurement matrix satisfies certain conditions such as the restricted isometry property (RIP)~\cite{blanchard2015performance}. All these methods consider the problem of recovering the unknown sparse signal using the available measurement vector and measurement matrix.
In sequential or online problems, the measurement vectors arrive sequentially and the learner predicts a sparse vector to best approximate the measurement vector, which is \textit{yet to arrive}. 
Because of this fundamental difference between the offline and online problems, the online problem will be henceforth referred to as the \textit{online sparse linear approximation (OSLA)}. Such online prediction problems arise frequently in diverse applications, such as medical trials~\cite{kale2017adaptive}, web caching~\cite{bhattacharjee2020fundamental,mukhopadhyay2021online}, resource allocation~\cite{chen2017}, etc.
A reasonable metric for evaluating the performance of an online policy for OSLA is the \textit{static regret}, defined as the difference between the cumulative prediction loss suffered by the learner and the optimal cumulative prediction loss calculated in hindsight by an omniscient adversary having perfect knowledge of the entire sequence. The learner for OSLA aims to develop an online policy that enjoys sublinear static regret.

The literature on OSLA is relatively recent. The papers~\cite{ziniel2013dynamic,asif2014sparse,vaswani2016recursive} address the dynamic compressed sensing (CS) which predominantly considers the recovery of a sparse vector from sequential linear measurements, often with additional assumptions on the evolution of the sparse vector. These works, however, consider time-varying sparse vector recovery, which observes $\bm{y}_t$ before estimating $\bm{x}_t$. The works~\cite{fosson2017online,fosson2020centralized} perform \textit{dynamic} regret analysis of online sparse recovery problem in the online convex optimization (OCO) framework which neglects non-convex sparsity constraints and considers a realizable model for the measurement sequence with bounded measurement noise. Another relevant area of research is the online sparse linear regression (OSLR). In OSLR, at time $t$, the learner predicts a sparse vector $\bm{x}_t$, following which the adversary reveals the test input-output pair $\bm{w}_t$ and $y_t$, so that the learner incurs the approximation loss $(y_t - \bm{w}_t^\top \bm{x}_t)^2$~\cite{foster2016online,kale2017adaptive,ito2017efficient,ito2018online}. Here the goal is to devise an efficient online policy to minimize the static regret.
The OSLR setup is closer in principle to the OSLA framework. However, OSLR considers scalar measurement sequence, whereas OSLA addresses vector sequence. Another related research area is that of online caching and online $k-$experts, which addresses online linear optimization (OLO) (unlike squared loss in OSLA) under sparsity constraint~\cite{wang2019online,bhattacharjee2020fundamental, mukhopadhyay2021online, mukhopadhyay2022k}. 

This letter serves as a pioneering work in OSLA by proposing a class of efficient online policies relevant to OSLA. We summarize below the salient contributions of this work:
\begin{itemize}
\item\textbf{Efficient online meta-policy for OSLA}: We design and analyze an efficient no-regret online meta-policy, referred to as \textit{Follow The Approximate Sparse Leader} (\texttt{FTASL}), which uses any efficient greedy sparse recovery algorithm to address the problem of OSLA. 
\item\textbf{Regret Analysis}: Under a certain \textit{realizability} assumption, we find an upper bound on the static regret of the proposed \texttt{FTASL} meta-policy. We show that \texttt{FTASL} enjoys variable regret guarantees, ranging from logarithmic to square-root, depending upon the specific structure of the problem. 
\end{itemize}
\textit{Organization of paper:} The remainder of the paper is organized as below. In Section~\ref{sec:online-sparse-approx} we describe the problem and develop the \texttt{FTASL} meta-policy for OSLA. Then theoretical analysis of the regret of \texttt{FTASL} is carried out in Section~\ref{sec:analysis}, followed by numerical simulations in Section~\ref{sec:numerical}. Finally, we conclude the paper in Section~\ref{sec:conclude}.\\
\textit{Preliminaries and Notations:} Throughout the paper, we use the notation $f(t) = \mathcal{O}(g(t))$ to denote that there exists $C\ge 0$ such that for all large $t$, $f(t)\le Cg(t)$. Similarly, we use $f(t) = \Theta(g(t))$ to denote that $\exists c,C\ge 0$ such that for all large $t$, $cg(t)\le f(t)\le Cg(t)$. We denote $[N] = \{1,\cdots, N\}.$ 
The support of a vector $\bm{x}\in \real^N$ is defined by the set $\{i\in [N]:x_i\ne 0\},$.
A matrix $\bm{\Phi}\in \real^{M\times N}$ is said to satisfy restricted isometry property (RIP) of order $K$ if the restricted isometry constant (RIC) of order $K$, defined by $\delta_K=\max_{S\subset [N]:\abs{S}=K}\opnorm{\bm{\Phi}_S^\top \bm{\Phi}_S - \bm{I}_K}{2\to 2}$, satisfies $\delta_K\in (0,1)$~\cite{foucart2011hard}. 
\section{Proposed Online Policy for OSLA}
\label{sec:online-sparse-approx}
%
%
\subsection{Problem formulation}
\label{sec:problem-formulation}
In OSLA, we are given a measurement matrix $\bm{\Phi}\in \real^{M\times N}$ with columns $\bm{\phi}_1,\cdots, \bm{\phi}_N$ along with sequentially arriving measurement vectors $\{\bm{y}_t\},\ \bm{y}_t \in \real^M$. Time is slotted and at each instant $t\ge 1$, the learner predicts a vector $\bvec{x}_t$ with support $S_t\subset [N]$ of size $K$. Following this, the nature reveals a vector $\bm{y}_t$ and the learner incurs a loss given by the approximation error, $\frac{1}{2}\norm{\bvec{y}_{t}-\bvec{\Phi x}_t}^2$. 
A standard metric to evaluate the performance of any online policy for OSLA is the 
\emph{static regret}, which compares the cumulative loss of the online policy with the minimum cumulative loss of the optimal offline static policy in hindsight. It is defined as follows:
\begin{align}
	\label{eq:regret-osa}
	R_{T} 
    \ & = \sum_{t=1}^T \frac{1}{2}\norm{\bvec{y}_t-\bvec{\Phi x}_t}^2 -\textrm{OPT},
\end{align}
where $\mathcal{B}^N_K = \{\bvec{x}\in \real^N:\opnorm{\bvec{x}}{0}=K\},$ and $\textrm{OPT} = \min_{\bvec{x}\in \mathcal{B}^N_K}\sum_{t=1}^T \frac{1}{2}\norm{\bvec{y}_t-\bvec{\Phi x}}^2.$  The goal is to obtain an online policy which enjoys \emph{sublinear} static regret.
\subsection{Algorithm Development}
\label{sec:algo-development}
%
%
\begin{algorithm}[t!]
\scriptsize
	\caption{\textsc{Follow-The-Approximately-Sparse-Leader \texttt{(FTASL)} Meta-Policy}}
	\label{algo:FTASL-meta}
	\begin{algorithmic}[1]
		\State {\bf Input:} $\mu>0\ K,\bvec{\Phi},\{\bm{y}_t\}_{t=1}^T$, \texttt{ALG}
		\State {\bf Initialize:} $\bvec{Y}_0=\bvec{0},\ \bm{b}_0 = \bm{0},\ \tau_0 = 0.$
		\For{$t=1,\cdots, T$}
        \State {\color{gray}\textit{\#\texttt{A-FTASL}}}
            \State $\bm{x}_t = 
		\mbox{\texttt{ALG}}(\bm{b}_{t-1},K,\bm{\Phi},\tau_{t-1}) $\
        \State {\color{gray}\textit{\#\texttt{L-FTASL}}}
		\If{$t$ is a power of $2$}
		\State $\bm{x}_t = 
		\mbox{\texttt{ALG}}(\bm{b}_{t-1},K,\bm{\Phi},\tau_{t-1}) $
			\Else
			\State $\bm{x}_t = \bm{x}_{t-1}.$
			\EndIf
			\State Reveal $\bvec{y}_t$ and suffer loss $\frac{1}{2}\norm{\bm{y}_t - \bm{\Phi x}_t}^2$ 
			\State Accumulate $\bvec{Y}_t = \bvec{Y}_{t-1} + \bvec{y}_t$
			\State $\bm{b}_{t} =\frac{\bm{Y}_{t}}{t},\ \tau_t = \lceil\ln (t+1)\rceil$
			\EndFor
		\end{algorithmic}
	\end{algorithm}
	%
	%
	To develop an online learning algorithm to address OSLA, we first observe that the optimal offline minimizer $\bm{x}_{\mathrm{opt}}$ of \textrm{OPT} can be expressed as below
	\begin{align}
    \label{eq:offline-optimum}
		\bm{x}_{\mathrm{opt}} = \argmin_{\bm{x}\in \mathcal{B}^N_K}\frac{1}{2}\norm{\frac{\bvec{Y}_{T}}{T} - \bm{\Phi x}}^2,
	\end{align}
	where $\bm{Y}_t = \sum_{\tau=1}^t \bm{y}_\tau,\ t\ge 1$. One could be tempted to define $\bm{x}_t=\argmin_{\bm{x}\in \mathcal{B}^N_K}\frac{1}{2}\norm{\frac{\bvec{Y}_{t}}{t} - \bm{\Phi x}}^2.$ However, such a policy is infeasible as $\bm{x}_t$ is predicted before the learner observes $\bm{y}_t$. {Follow-The-Leader} (\texttt{FTL})~\cite{kalai2005efficient} attempts to circumvent this difficulty by proposing the following policy:
	\begin{align}
		\label{eq:inefficient-online-policy-ftl}
		\bm{x}^\star_t & = \argmin_{\bvec{x}\in \mathcal{B}^N_K}\norm{\bm{b}_{t-1} - \bm{\Phi x}}^2,\ t\ge 1,
	\end{align}
	where $\bm{b}_t=\left\{\begin{array}{cc}
		\bm{0}, & t=0  \\
		\frac{\bm{Y}_{t}}{t}, & t\ge 1. 
	\end{array}\right.$
	Unfortunately, the optimization problem in~\eqref{eq:inefficient-online-policy-ftl} is well-known to be NP-Hard~\cite{davis1997adaptive}, making the above policy inefficient. 
	
	To avoid the aforementioned difficulty, we propose approximating~\eqref{eq:inefficient-online-policy-ftl} by running $\tau_t$ iterations of an efficient greedy sparse recovery algorithm, referred to as \texttt{ALG}. We denote by $\bm{x}_t = \texttt{ALG}(\bm{b}_{t-1},K,\bm{\Phi},\tau_t)$ the estimate of $\bm{x}_t^\star$ produced by \texttt{ALG}. The resulting online meta-policy is referred to as Follow The Approximate Sparse Leader (\texttt{FTASL}) and is described in Algorithm~\ref{algo:FTASL-meta}. Depending on the frequency of updates of $\bm{x}_t$, we propose two versions of the \texttt{FTASL}, the \textit{agile} and the \textit{lazy} versions, referred to as the \texttt{A-FTASL} and \texttt{L-FTASL}, respectively. In the agile version, $\bm{x}_t$ is updated each time $t$, whereas, in the lazy version, the updates take place only when $t$ is a power of $2$. We will see that agile version has sublinear worst-case regret guarantee, with linear computational complexity. On the other hand, the lazy updates reduce the computational complexity drastically, albeit compromising the regret guarantee, which becomes sublinear under special circumstances. 
\label{sec:fthtpl}
%

\section{Performance Analysis}
\label{sec:analysis}
In this section, we analyze the regrets as well as the computational complexities of the proposed \texttt{A-FTASL} and \texttt{L-FTASL}.
\paragraph{Regret Analysis}
\label{sec:regret-analysis}
To analyze the static regret of the proposed \texttt{FTASL} policy, we adopt the following assumptions:\\
\emph{Assumption 1} (\textbf{Realizability}): 
    Each $\bm{y}_t$ satisfies
\begin{align}
		\label{eq:yt-model-assumption}
		\bm{y}_t & = \bm{\Phi u}_t + \bm{w}_t,\ t\ge 1,
	\end{align}
	where $\{\bm{u}_t\}$ is a sequence of $K-$sparse vectors with a \emph{common} support $\Lambda$ and with bounded norms $\norm{\bm{u}_t}\le 1$. Furthermore, the measurement errors are i.i.d. Gaussian, i.e., for all $t$, $\bm{w}_t\sim\mathcal{N}(\bm{0},\bm{I})$.\\
\emph{Assumption 2} (\textbf{Algorithmic stability}): If $\bm{b}=\bm{Au}+\bm{e}$, for some $\bm{A},\bm{u},\bm{e}$, and if  \texttt{ALG} produces an estimate $\hat{\bm{u}}$ after running for $\tau$ iterations, then there is $\kappa>0$ such that the following guarantee holds:
	\begin{align}
		\label{eq:alg-stability-assumption}
		\norm{\hat{\bm{u}}-\bm{u}} & \le 2^{-\tau} \norm{\bm{u}} + \kappa\norm{\bm{e}}.
	\end{align}
The literature shows that such guarantees can be ensured by requiring that the measurement matrix satisfies certain RIP, such as, $\delta_{3K}<0.618$ for IHT~\cite{zhao2023improved}, $\delta_{3K}<\frac{1}{\sqrt{3}}$ for HTP~\cite{foucart2011hard}, etc. 

\noindent\textbf{Regret Decomposition}:
We begin the regret analysis with the following decomposition: 
\begin{align}
	R_T & = \underbrace{\sum_{t=1}^T \frac{1}{2}\norm{\bm{y}_t - \bm{\Phi z}^\star_{T}}^2 - \min_{\bm{x}\in \mathcal{B}^N_K}\sum_{t=1}^T \frac{1}{2}\norm{\bm{y}_t - \bm{\Phi x}}^2}_{A}\nonumber\\
	\ & + \underbrace{\sum_{t=1}^T \left[\frac{1}{2}\norm{\bm{y}_t - \bm{\Phi z}^\star_t}^2 - \frac{1}{2}\norm{\bm{y}_t - \bm{\Phi z}_{T}^\star}^2\right]}_{B}\nonumber\\
	\label{eq:regret-decomposition}
	\ & + \underbrace{\sum_{t=1}^T \left[\frac{1}{2}\norm{\bm{y}_t - \bm{\Phi x}_t}^2 - \frac{1}{2}\norm{\bm{y}_t - \bm{\Phi z}_{t}^\star}^2\right]}_{C},
\end{align}
where $\bm{z}_t^\star = \frac{\sum_{s=1}^t \bm{u}_s}{t},\ t\ge 1$. Due to the involvement of the measurement noise sequence $\{\bm{w}_t\}_{t=1}^T$, all the terms $A,B,C$ above are random. We obtain high probability bounds on each of these terms, stated in the result below: 
    \begin{lem}
        \label{lem:high-prob-bounds-A-B-C}
        Let $\delta\in (0,1)$ and define $b(\delta) = (\sqrt{M}+\sqrt{\ln(1/\delta)}^2+\ln(1/\delta).$ Then,
        \begin{align}
            \label{eq:A-upper-bound-lem}
            A & \le \frac{b(\delta)}{2},\ \mathrm{w.p.}\ge 1-\delta,\\
            \label{eq:B-upper-bound-lem}
            B &\le \sqrt{2(1+\delta_K)\ln(1/\delta)\sum_{t=1}^T\norm{\bm{z}_T^\star - \bm{z}_t^\star}^2} \  \mathrm{w.p.}\ge 1-\delta,\\
		C & \le \sqrt{2\ln(2/\delta)(1+\delta_K)a_T(\delta)} + \frac{a_T(\delta)(1+\delta_K)}{2}\nonumber\\ 
                    \label{eq:C-upper-bound-lem}
\ & + (1+\delta_K)\sqrt{a_T(\delta)\sum_{t=1}^T \norm{\bm{z}_t^\star-\bm{u}_t}^2} \mathrm{w.p.}\ge 1-\delta,
	\end{align}
    where
    \begin{align}
    \label{eq:aTdelta)}
        a_T(\delta) & = 1+2\kappa^2 Mb(\delta)\ln T +\Delta,\\
        \Delta & =2\sum_{k=0}^{k'-1}\Delta_k,\\
    \Delta_k & = \left\{\begin{array}{ll}
         0, &  \mbox{\texttt{A-FTASL}}\\ \sum_{t=t_k}^{t_{k+1}-1}\norm{\bm{z}_t^\star-\bm{z}_{t_k-1}^\star}^2, &  \mbox{\texttt{L-FTASL}}
        \end{array}
\right.
    \end{align}
\end{lem}
and $t_k = 2^k,k\ge 0$ with $k'$ an integer such that $t_{k'-1}\le T\le t_{k'}-1.$    
	The derivation of the bounds~\eqref{eq:A-upper-bound-lem},\eqref{eq:B-upper-bound-lem} and~\eqref{eq:C-upper-bound-lem} requires careful analysis of the way the noise sequence $\{\bm{w}_t\}$ interacts in each of these terms. The details of the analysis can be found in Appendix~\ref{sec:appendix-proof-lem-high-prob-bounds-a-b-c}. With these bounds, we now state the main result of this paper:
    \begin{theorem}
        \label{thm:regret-bound}
        Let the \textbf{realizability} and \textbf{algorithmic stability} assumptions hold. Then, for any $\delta\in (0,1)$, the following holds with probability $\ge 1-\delta$:
        \begin{align}
		\lefteqn{R_T \le \frac{b(\delta/3)}{2} + \sqrt{2(1+\delta_K)\ln(3/\delta)\sum_{t=1}^T\norm{\bm{z}_T^\star - \bm{z}_t^\star}^2}} & &\nonumber\\
		\ & + \sqrt{2\ln(6/\delta)(1+\delta_K)\left(1+2M\kappa^2b(\delta/3)\ln T+\Delta\right)}\nonumber\\
		\ & + \frac{(1+\delta_K)}{2}\left(12M\kappa^2b(\delta/3)\ln T+\Delta\right)\nonumber\\
        \label{eq:R_T_bound}
		\ & +(1+\delta_K)\sqrt{\left(1+2M\kappa^2b(\delta/3)\ln T+\Delta\right)\sum_{t=1}^T \norm{\bm{z}_t^\star-\bm{u}_t}^2}.
	\end{align}
    \end{theorem}
    \begin{proof}
        Let us denote the right hand side (RHS) of~\eqref{eq:R_T_bound}as $\textrm{RHS}(\delta)$ (to show its dependence on $\delta$) and the RHSs of~\eqref{eq:A-upper-bound-lem},~\eqref{eq:B-upper-bound-lem} and~\eqref{eq:C-upper-bound-lem} by $\textrm{RHS}_1(\delta)$,$\textrm{RHS}_2(\delta)$, and $\textrm{RHS}_3(\delta)$, respectively. Then we observe that
        \begin{align}
            \ & \mathbb{P}\left(R_T\le \textrm{RHS}(\delta)\right)\nonumber\\
            \ge & \mathbb{P}\left(A\le \textrm{RHS}_1(\delta/3),\ B\le \textrm{RHS}_2(\delta/3),\ C\le \textrm{RHS}_3(\delta/3)\right)\nonumber\\
            \stackrel{{\color{blue}\psi_1}}{\ge} & 1-\mathbb{P}\left(A> \textrm{RHS}_1(\delta/3)\right) -\mathbb{P}\left(B> \textrm{RHS}_2(\delta/3)\right)\nonumber\\
            - &
            \mathbb{P}\left(C> \textrm{RHS}_3(\delta/3)\right)\nonumber\\
            \stackrel{{\color{blue}\psi_2}}{\ge} & 1 - \frac{\delta}{3}-\frac{\delta}{3}-\frac{\delta}{3}=1-\delta,
        \end{align}
        where the step {\color{blue}{$\psi_1$}} is due to the union bound and the step {\color{blue}{$\psi_2$}} is due to the inequalities~\eqref{eq:A-upper-bound-lem},\eqref{eq:B-upper-bound-lem},\eqref{eq:C-upper-bound-lem}.
    \end{proof}
    \noindent\textbf{Discussion}: There are several interesting implications of Theorem~\ref{thm:regret-bound}, which we discuss below:
	\begin{enumerate}
		\item When $\bm{u}_t = \bm{u},\ 1\le t\le T$, we have $\bm{z}_t^\star = \bm{u}$ for all $1\le t\le T$. Therefore, in this case, $\Delta_k=0,\ \forall k$, for \texttt{L-FTASL}, so that $\Delta=0$ for both \texttt{A-FTASL} and \texttt{L-FTASL}. Then~\eqref{eq:R_T_bound} implies that, w.p.$\ge 1-\delta$, for each of \texttt{L-FTASl} and \texttt{A-FTASL}, $R_T =
\mathcal{O}\left((1+\delta_K)M\kappa^2b(\delta/3)\ln T\right).$
		\item When there is no additional assumptions on $\{\bm{u}_t\}$, we have that $\norm{\bm{z}_t^\star - \bm{u}_t}\le 2$ and $\norm{\bm{z}_t^\star - \bm{z}_T^\star}\le 2$. Then, it follows from Eq.~\eqref{eq:R_T_bound} that, for \texttt{A-FTASL}, w.p.$\ge 1-\delta$, $R_T = \mathcal{O}\left((1+\delta_K)\sqrt{M\kappa^2b(\delta/3)T\ln T}\right)$.
        However, for \texttt{L-FTASL}, one can verify that $\Delta=\mathcal{O}(T)$, implying that, without additional assumptions, $R_T=\mathcal{O}(T)$ for \texttt{L-FTASL}.
        \item If there is a $\bm{z}$ such that $\norm{\bm{z}_t^\star-\bm{z}}=\mathcal{O}(1/\sqrt{t})$, then, $\Delta= \sum_{k=0}^{k'-1}t_k\times \mathcal{O}(1/t_k)=\mathcal{O}(k')=\mathcal{O}(\ln T).$ Then it follows from~\eqref{eq:R_T_bound} that for both \texttt{L-FTASL} and \texttt{A-FTASL}, w.p.$\ge 1-\delta$, $R_T = \mathcal{O}((1+\delta_K)\sqrt{M\kappa^2b(\delta/3)T\ln T})$. 
	\end{enumerate}
    \paragraph{Computational Complexity Analysis}
    \label{sec:computational-complexity}
    In this section, we analyze the computational complexities of the \texttt{A-FTASL} and \texttt{L-FTASL} algorithms and emphasize on their dependence on time. Note that when \texttt{FTASL} updates $\bm{x}_t$, it has to execute $\tau_t$ iterations of \texttt{ALG}. Assuming that \texttt{ALG} has a per-iteration computational complexity of $C$ units ($C$ obviously depends on the problem parameters such as $M,K$ etc), the computational complexity of an iteration of \texttt{FTASL}  when it updates is $C\tau_t$. Hence, for \texttt{A-FTASL}, the total computational complexity is $\sum_{t=1}^T C\tau_t\le \sum_{t=1}^T C(1+\ln(t+1))=\mathcal{O}(T\ln T)$. For \texttt{L-FTASL}, the complexity becomes, $\sum_{k=0}^{k'-1}C\ln(2^k+1)=\mathcal{O}((\ln T)^2).$
\section{Numerical Results}
\label{sec:numerical}
%
%
In this section, we evaluate the performances of the proposed \texttt{FTASL} meta-algorithm, both the agile and lazy versions, with \texttt{OIST}~\cite{fosson2017online} and \texttt{ODR}~\cite{fosson2020centralized} as the benchmarks. We use the IHT~\cite{blumensath08iht} and HTP~\cite{foucart2011hard} algorithms as the candidates for \texttt{ALG}, both enjoying provable recovery guarantees~\cite{blanchard2015performance}~\footnote{It should be noted that we can use other greedy algorithms with provable recovery guarantees too, such as CoSAMP~\cite{needell2009cosamp} and SP~\cite{dai2009subspace} }. For simulations, we select the entries of $\bm{\Phi}$ as i.i.d.$\sim\mathcal{N}(0,1/\sqrt{M})$, which ensures that $\bm{\Phi}$ satisfies the RIP with high probability~\cite{foucart2013}. In all the experiments the $\{\bm{w}_t\}$ sequence is sampled from i.i.d. $\mathcal{N}(\bm{0},\bm{I}_M)$. Following~\cite{fosson2017online}, we choose the OIST step size to be $0.02\opnorm{\bm{\Phi}}{2\to2}^{-2}$. Furthermore, for both \texttt{OIST} and \texttt{ODR}, we choose the parameters $\lambda=0.01$ with $r = 14$~\cite[\S 7]{fosson2020centralized}.
\paragraph{Regret performance}
\begin{figure}[t!]
	\centering
	\begin{subfigure}{0.24\textwidth}
		\centering
		\includegraphics[scale=0.26]{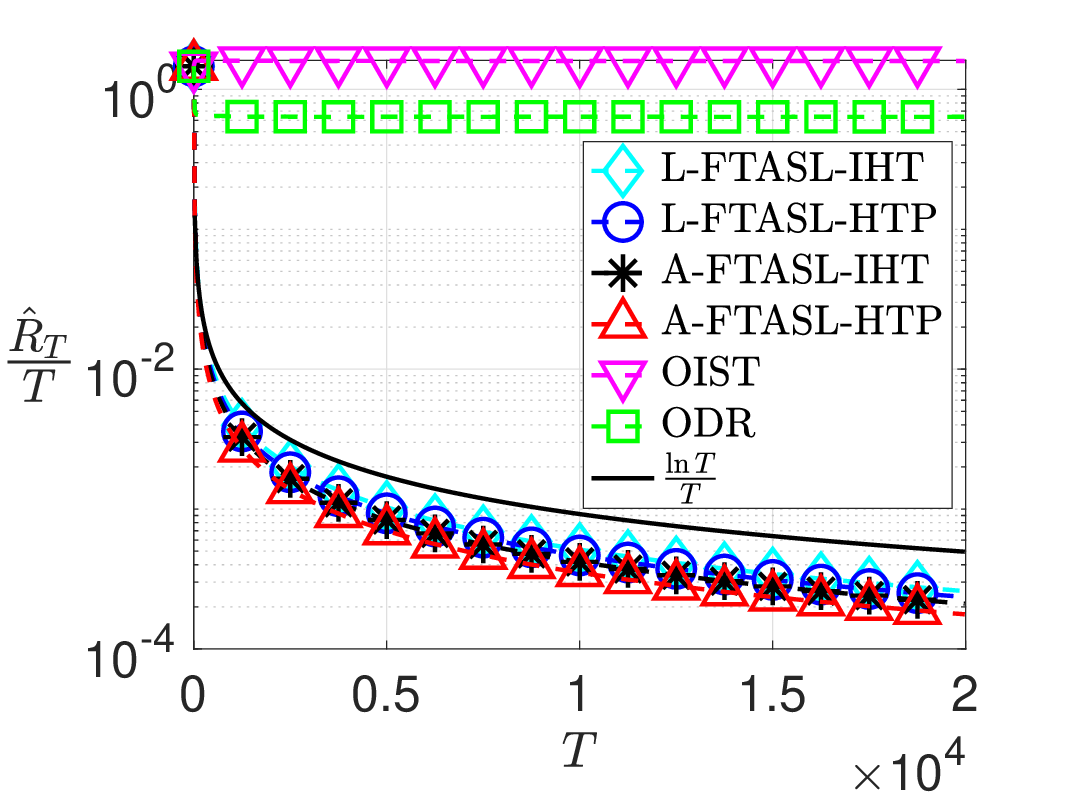}
		\subcaption{}
	\end{subfigure}
	\begin{subfigure}{0.24\textwidth}
		\centering
		\includegraphics[scale=0.26]{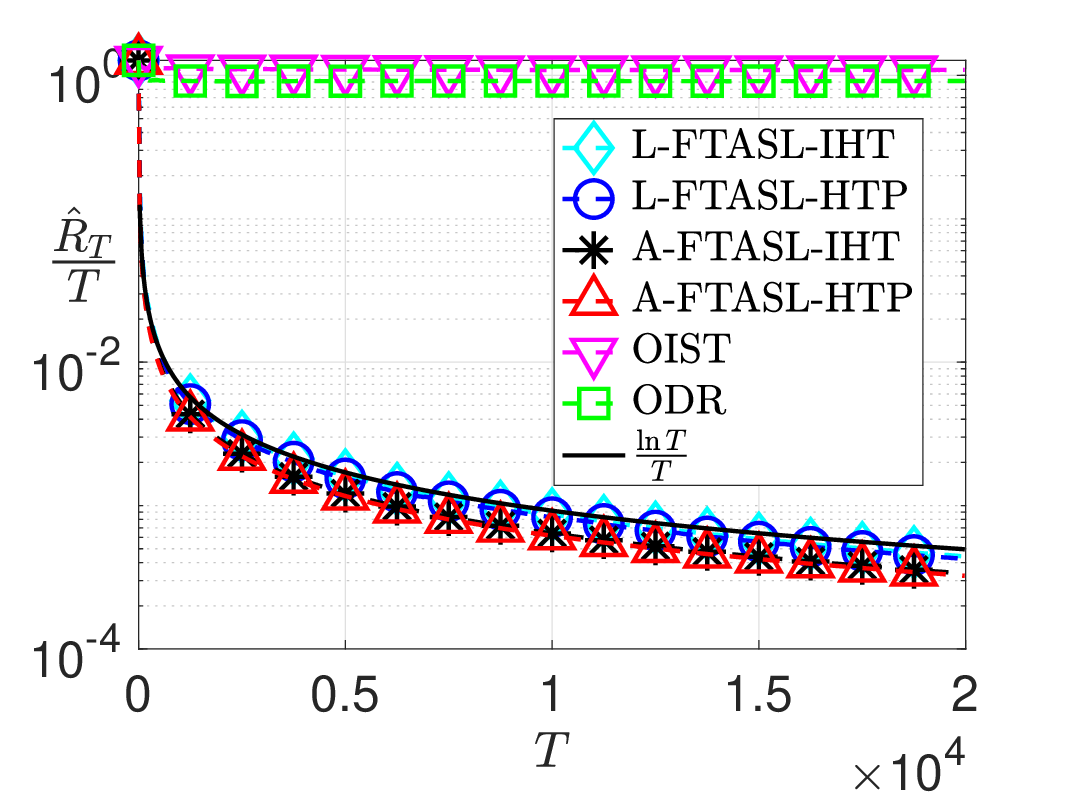}
		\subcaption{}
	\end{subfigure}
    \begin{subfigure}{0.24\textwidth}
		\centering
		\includegraphics[scale=0.26]{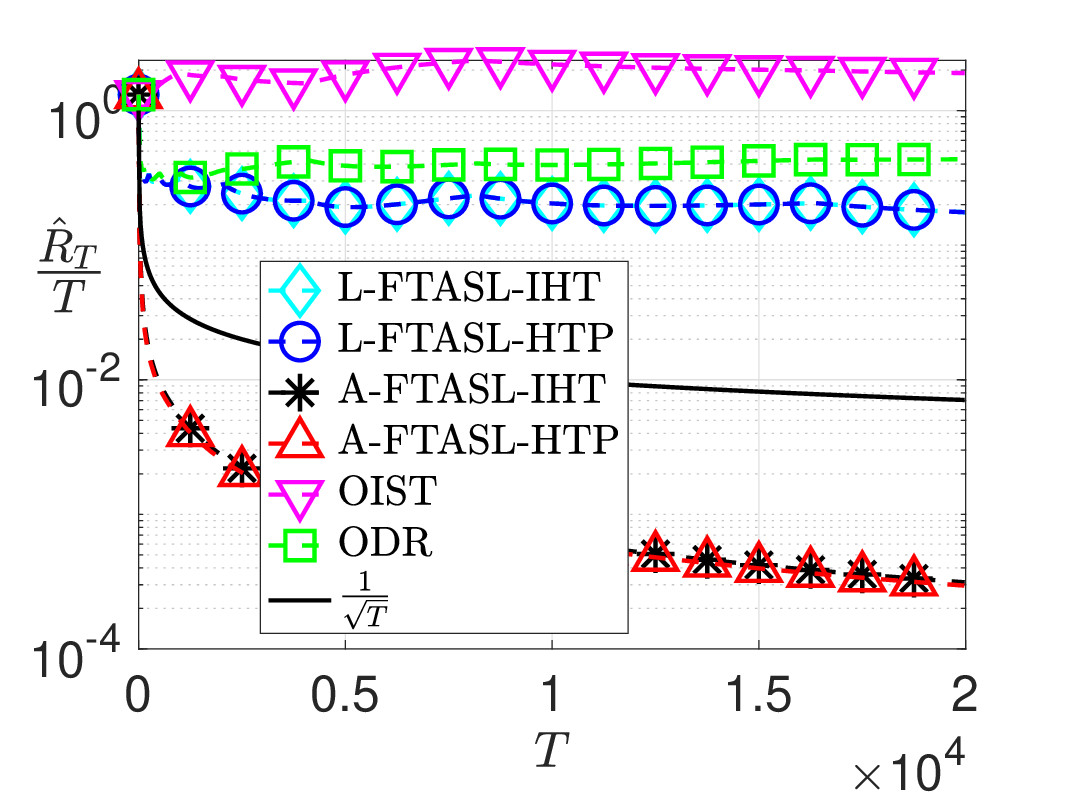}
		\subcaption{}
	\end{subfigure}
    \begin{subfigure}{0.24\textwidth}
		\centering
		\includegraphics[scale=0.26]{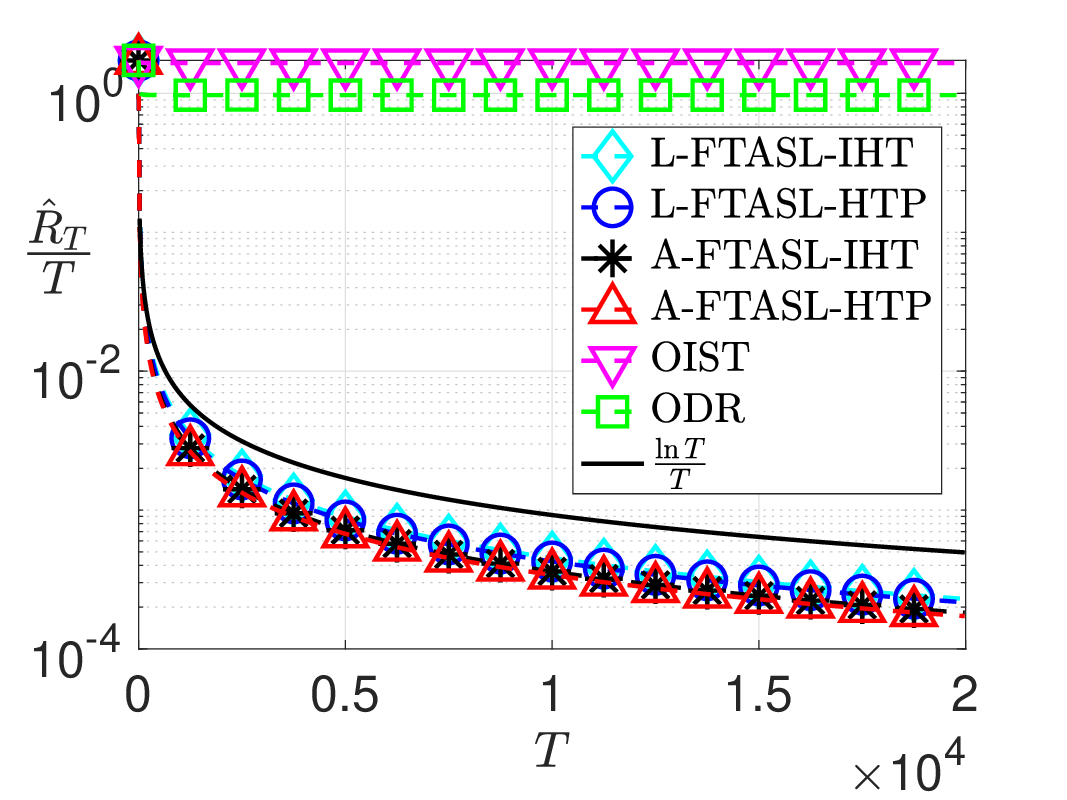}
		\subcaption{}
	\end{subfigure}
    \caption{\footnotesize{Time-averaged regret vs time (a) for synthetic data with $\bm{u}_t = \bm{u}$, (b) for synthetic data with variable $\bm{u}_t$ such that $\bm{u}_{t,i}\sim \mathcal{U}[0,1]$ i.i.d. $\forall i\in \Lambda$, with $M=256,N=512,K=10$, (c) for synthetic data with variable $\bm{u}_t$ such that $u_{t,i}\sim \mathcal{U}[0,1]$ is updated if $t$ is a power of $2$, (d) for ensemble average over different DIGITS figures with $N=784,\ M=392,\ K=10$.}} 
    \label{fig:regret}
\end{figure}
Since computing \textrm{OPT} in~\eqref{eq:offline-optimum} is difficult, in general, we use the following approximation of $\textrm{OPT}$ to compute an approximate regret $\hat{R}_T$ as below:
\begin{align}
    \label{eq:approx-regret}
    \hat{R}_T = \sum_{t=1}^T\frac{1}{2}\norm{\bm{y}_t - \bm{\Phi x}_t}^2 - \sum_{t=1}^T\frac{1}{2}\norm{\bm{y}_t - \bm{\Phi z}_T^\star}^2. 
\end{align}
From the regret decomposition in Eq.~\eqref{eq:regret-decomposition}, observe that $R_T - \hat{R}_T = A$. It can be shown that with high probability, $\hat{R}_T = \Theta(R_T)$.~\footnote{To see this note that as $\bm{z}_T^\star$ is $K-$sparse, we have $\hat{R}_T\le R_T$; and from Eq.~\eqref{eq:A-upper-bound-lem}, with probability $\ge 1-\delta$, $R_T - \hat{R}_T \le \frac{b(\delta)}{2}.$} 
%
%
%
%
For synthetic data, we take $M = 256, N=512, K=10$ with $\bm{u}_t = \bm{u}$, for all $1\le t\le T$, where the support $\Lambda$ of size $K$ chosen uniformly randomly from $[N]$, and the non-zero entries of $\bm{u}$ are chosen i.i.d.$\sim \mathcal{U}[0,1]$. In Fig.~\ref{fig:regret}(a), we plot $\hat{R}_T/T$, with respect to $T$. The plot reveals that $\hat{R}_T/T$ is constant for \texttt{OIST} and \texttt{ODR}, implying that these policies suffer from linear regret. In contrast, both the \texttt{A-FTASL} as well as \texttt{L-FTASL} meta-policies enjoy logarithmic regret for either of IHT or HTP as \texttt{ALG}. This matches with the theoretical guarantee derived from Theorem~\ref{thm:regret-bound}.
In Fig.~\ref{fig:regret}(b), $\hat{R}_T/T$ is plotted against $T$ when the non-zero entries of $\bm{u}_t$ (i.e. $u_{t,i},\ i\in \Lambda$) are sampled i.i.d~$\sim\mathcal{U}[0,1]$ at each time. We observe that both \texttt{OIST} and \texttt{ODR} exhibit linear regret while both \texttt{A-FTASL} as well as \texttt{L-ALG} enjoy logarithmic regret. The latter can be explained by using Theorem~\ref{thm:regret-bound} along with the observation that the i.i.d. bounded sequence $\{\bm{u}_{t,\Lambda}\}_t$ is concentrated around $\mathbb{E}(\bm{u}_{t,\Lambda})=\frac{\bm{1}_{\Lambda}}{2}$ with high probability.~\footnote{This is a consequence of concentration inequalities, e.g., Hoeffding's inequality~\cite{boucheron2003concentration}} 
In Fig.~\ref{fig:regret}(c), again $\hat{R}_T/T$ is plotted against $T$ a time-varying $\{\bm{u}_t\}$ where the non-zero entries of $\bm{u}_t$ are updated only when $t$ is a power of $2$. The plots reveal that while \texttt{OIST}, \texttt{ODR} as well as \texttt{L-FTASL} suffer from linear regret only \texttt{A-FTASL} enjoys sublinear regret. This can be explained by Theorem~\ref{thm:regret-bound} by observing that the sequence $\{\bm{u}_t\}$ under consideration forces $\Delta=\mathcal{O}(T)$ for \texttt{L-FTASL}. 
%
%
%
\begin{figure}[t!]
    \centering
    \includegraphics[scale=0.26]{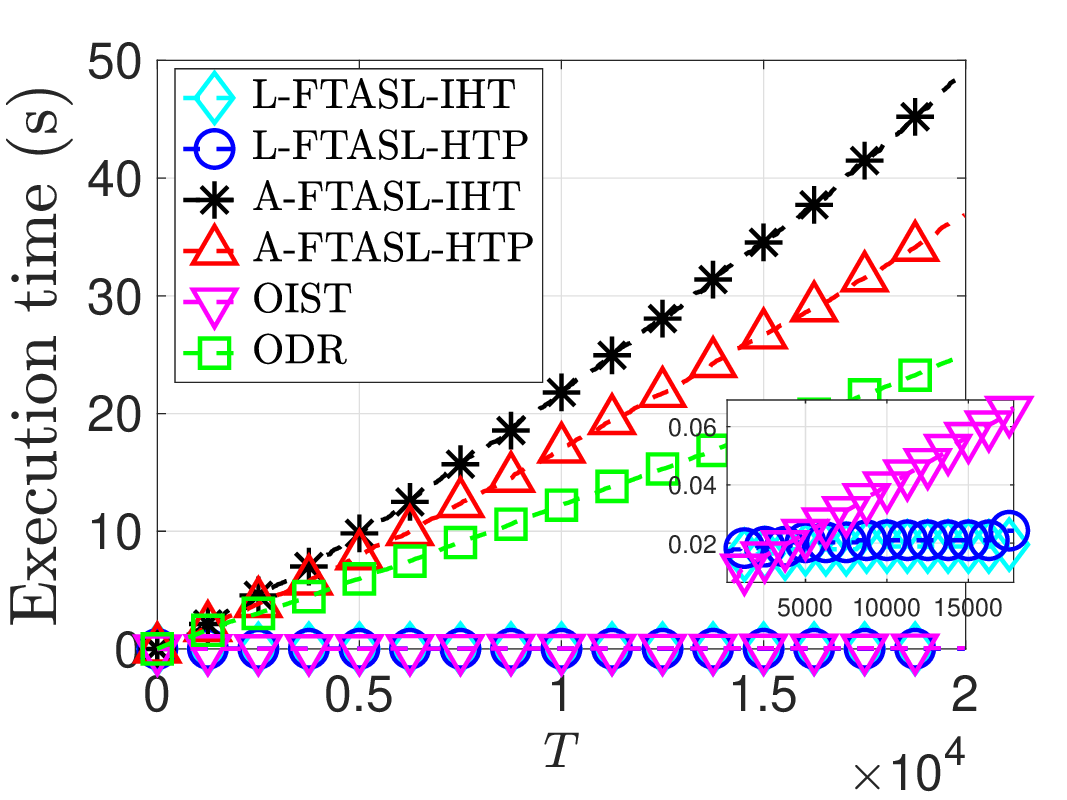}
    \caption{\footnotesize{Execution time vs $T$.}}
    \label{fig:execution-time}
\end{figure}

For testing the performance of the proposed online policies on real datasets, we use the DIGITS dataset from the UCI machine learning repository~\cite{alpaydin1998pen} which consists of $10992$ images of handwritten digits, each of size $28\times 28$. 
Each pixel value is in the range $[0,1]$ and we sparsify the images by retaining only the pixel values higher than $0.95$. The resulting images are converted into vectors of length $N=784$. We take $M =392$ and $K=10$, so that we only consider those processed images for which the sparsity is smaller than $10$. 
For this experiment, we run $10$ independent trials, where in each trial, the sequence $\{\bm{u}_t\}$ is taken as a constant vector constructed from one of the processed DIGITS images, sampled uniformly randomly from the processed dataset. Then we find the ensemble average of the time-average static regrets. The performance of the \texttt{FTASL}-based policies along with \texttt{OIST} and \texttt{ODR} are evaluated and plotted in Fig.~\ref{fig:regret}(c). We again observe that the policies \texttt{OIST} and \texttt{ODR} exhibit linear regret, while the proposed \texttt{A-FTASL} and \texttt{L-FTASL} policies exhibit logarithmic regret.    
\paragraph{Execution time performance}
We compare the execution time performance of the \texttt{FTASL} variants, using IHT and HTP as \texttt{ALG}, with \texttt{OIST} and \texttt{ODR} as benchmarks. For this experiment we use the setting of Fig.~\ref{fig:regret}(a), i.e., $\bm{u}_t$ is fixed for all time. In Fig.~\ref{fig:execution-time} we plot the execution time of the different algorithms for running it for $T$ iterations vs $T$. From the figure we see that while the executions times of \texttt{ODR}, \texttt{OIST} and \texttt{A-FTASL} increase linearly, the execution time of \texttt{L-FTASL} is almost constant. This observation matches well with the complexity analysis of $\mathcal{O}(T\ln T)$ for \texttt{A-FTASL} and $\mathcal{O}((\ln T)^2)$ for \texttt{L-FTASL}.
%
	
%
%
\section{Conclusion and Future Work}

\label{sec:conclude}
In this letter, we have proposed a greedy sparse recovery assisted \texttt{FTL} based online meta-policy for the OSLA problem, termed as \texttt{FTASL},  which uses enjoys $\mathcal{O}\left(\ln T\right)$ regret for a certain asymptotically realizable model, both proven theoretically as well as verified numerically. 
Potential future works consist of extending this result for time varying sensing matrices, and for general adversarial measurement sequences.
\appendix
\section{Proof of Lemma~\ref{lem:high-prob-bounds-A-B-C}}
\label{sec:appendix-proof-lem-high-prob-bounds-a-b-c}
\subsection{Upper bounding $A$:}
\label{sec:upper-bounding-a}
	%
	%
	First note that $ \bm{x}_{\mathrm{opt}} =\argmin_{\bm{x}\in \mathcal{B}^N_K}\sum_{t=1}^T \frac{1}{2}\norm{\bm{y}_t - \bm{\Phi x}}^2$ $= \argmin_{\bm{x}\in \mathcal{B}^N_K}\norm{\frac{\bm{Y}_T}{T}-\bm{\Phi x}}^2.$
	This allows one to rewrite $A$ as below:
	\begin{align}
		\label{eq:A-rewrite}
		A & = \frac{T}{2}\left[\norm{\frac{\bm{Y}_T}{T}-\bm{\Phi z}_T^\star}^2 - \norm{\frac{\bm{Y}_T}{T}-\bm{\Phi x}_{\mathrm{opt}}}^2\right].
	\end{align}
	
	Denote $\bm{h} = \bm{x}_{\mathrm{opt}} - \bm{z}_T^\star$. Then, one can express $A$ as below:
	\begin{align*}
		A & = \frac{T}{2}\left[\norm{\frac{\bm{Y}_T}{T}-\bm{\Phi z}_T^\star}^2 - \norm{\frac{\bm{Y}_T}{T}-\bm{\Phi z}_{T}^\star-\bm{\Phi h}}^2\right]\\
		\ & = \frac{T}{2}\left[\norm{\bm{\nu}_T}^2 - \norm{\bm{\nu}_T-\bm{\Phi h}}^2\right],
	\end{align*}
	where we use the relation $\frac{\bm{Y}_T}{T}-\bm{\Phi z}_T^\star = \frac{1}{T}\sum_{t=1}^T \bm{w}_t=:\bm{\nu}_T.$
	Let $\Gamma = \Lambda \cup S$, where $S$ is the support of $\bm{x}_{\mathrm{opt}}$. Then $\bm{h}$ is supported on $\Gamma$. Clearly, the minimizer $\bm{v}$ of $\frac{1}{2}\norm{\bm{\nu}_T - \bm{\Phi v}}^2$, such that $\bm{v}$ is supported on $\Gamma$ is $\bm{v}^\star$, where $\bm{v}^\star_\Gamma = \bm{\Phi}_\Gamma^\dagger\bm{\nu}_T$. Here, we have assumed that $\bm{\Phi}$ satisfies the RIP of order $2K$, since $\Gamma$ is of size at most $2K$. Consequently, we obtain, 
	\begin{align}
		A & \le \frac{T}{2}\norm{\proj{\Gamma}\bm{\nu}_T}^2\le \frac{T}{2}\norm{\bm{\nu}_T}^2.
	\end{align}  
	Since $\bm{w}_t\sim \mathcal{N}(\bm{0},\bm{I})$, $\bm{\nu}_T$ is a Gaussian vector. Furthermore, Since each of $\bm{w}_t\sim \mathcal{N}(\bm{0},\bm{I}_M)$ and the sequence $\{\bm{w}_t\}$ is i.i.d., we obtain, 
	\begin{align*}
		\mathbb{E}(\bm{\nu}_T) & = \frac{\sum_{t=1}^T \mathbb{E}(\bm{w}_t)}{T} = \bm{0},\nonumber\\
		\mathbb{E}(\bm{\nu}_T \bm{\nu}_T^\top) & = \frac{\sum_{t,\tau=1}^T\mathbb{E}(\bm{w}\bm{w}_{\tau}^\top)}{T^2} = \frac{\sum_{t=1}^T \mathbb{E}(\bm{w}_t\bm{w}_t^\top)}{T^2} = \frac{\bm{I}_M}{T}.
	\end{align*} Therefore $\bm{\nu}_T\sim \mathcal{N}\left(\bm{0}, \frac{1}{T}\bm{I}_M\right),$ so that $T\norm{\bm{\nu}_T}^2$ is distributed as a $\chi^2$ random variable with $M$ degrees of freedom. Using the concentration inequality for central $\chi^2$ random variables from~\cite{laurent2000adaptive}, we obtain, for any $\delta>0$, with probability $\ge 1-\delta$ the following holds:
	\begin{align}
		\label{eq:chi-square-concentration}
		T\norm{\bm{\nu}_T}^2 \le (\sqrt{M}+\sqrt{\ln(1/\delta)})^2+\ln(1/\delta).
	\end{align} 
    This results in the bound in Eq.~\eqref{eq:A-upper-bound-lem}.
	\subsection{Upper bounding $B$:}
    \label{sec:upper-bounding-b}
	Note that using the measurement model~\eqref{eq:yt-model-assumption} we can write, 
	\begin{align}
		B & = \sum_{t=1}^T\left[\frac{1}{2}\norm{\bm{\Phi}(\bm{u}_t - \bm{z}^\star_t)}^2 - \frac{1}{2}\norm{\bm{\Phi}(\bm{u}_t - \bm{z}^\star_T)}^2 \right]\nonumber\\
		\label{eq:D-bound}
		\ & + \sum_{t=1}^T\bm{w}_t^\top \bm{\Phi}(\bm{z}_T^\star - \bm{z}_t^\star).
	\end{align}
	Now observe that,
	\begin{align*}
		\ & \argmin_{\bm{z}\in \mathcal{B}^N_K}\sum_{\tau=1}^t \frac{1}{2}\norm{\bm{\Phi}(\bm{u}_{\tau} - \bm{z})}^2\\
		\ & =  \argmin_{\bm{z}\in \mathcal{B}^N_K}\frac{1}{2}\norm{\bm{\Phi}\left(\frac{\sum_{\tau=1}^t\bm{u}_{\tau}}{t} - \bm{z}\right)}^2\\
		\ & = \argmin_{\bm{z}\in \mathcal{B}^N_K}\frac{1}{2}\norm{\bm{\Phi}\left(\bm{z}_t^\star - \bm{z}\right)}^2.
	\end{align*}
	Now as $\bm{\Phi}$ satisfies RIP of order $2K$, the above minimum is attained uniquely at $\bm{z}_t^\star$.
	To proceed with, first consider the online optimization problem where at time $t$ the learner predicts $\bm{z}^\star_t$ to receive a loss of $\frac{1}{2}\norm{\bm{\Phi}(\bm{u}_t - \bm{z}^\star_t)}^2.$ The corresponding regret is given by the following:
	\begin{align*}
		\widetilde{R}_T & = \sum_{t=1}^T \frac{1}{2}\norm{\bm{\Phi}(\bm{u}_t - \bm{z}^\star_t)}^2 - \min_{\bm{z}\in \mathcal{B}_K^N}\sum_{t=1}^T \frac{1}{2}\norm{\bm{\Phi}(\bm{u}_t - \bm{z})}^2.
	\end{align*}
	Observe that 
	\begin{align}
		\bm{z}_t^\star = \argmin_{\bm{z}\in \mathcal{B}_K^N}\frac{1}{2}\norm{\bm{\Phi}(\bm{u}_t - \bm{z})}^2.
	\end{align} 
	Now using the measurement model~\eqref{eq:yt-model-assumption}, one obtains, for any $t\ge1$, $\bm{z}_t^\star = \frac{\sum_{s=1}^t \bm{u}_s}{t}= \argmin_{\bm{x}\in \mathcal{B}^N_K}\sum_{s=1}^t \frac{1}{2}\norm{\bm{\Phi}(\bm{u}_t - \bm{x})}^2$. It then follows from the well-known Be-The-Leader (BTL) argument~\cite[\S3]{kalai2005efficient} that the first summand of $B$ in~\eqref{eq:D-bound} is non-positive. Consequently, 
	\begin{align}
		\label{eq:B-rewrite}
		B & \le \sum_{t=1}^T\bm{w}_t^\top \bm{\Phi}(\bm{z}_T^\star - \bm{z}_t^\star).
	\end{align}
	Since $\{\bm{\Phi}(\bm{z}_T^\star - \bm{z}_t^\star)\}$ is a deterministic sequence, $\sum_{t=1}^T\bm{w}_t^\top \bm{\Phi}(\bm{z}_T^\star - \bm{z}_t^\star)$ is Gaussian with $0$ mean and variance $\sum_{t=1}^T \norm{\bm{\Phi}(\bm{z}_T^\star - \bm{z}_t^\star)}^2$. Since all the random variables $\bm{z}_t^\star, t=1,\cdots, T$, have a common support $\Lambda$, using the RIP we can further upper bound the variance as $(1+\delta_K)\sum_{t=1}^T \norm{\bm{z}_T^\star - \bm{z}_t^\star}^2$. Using the Chernoff bound for standard Gaussian random variables, we obtain the bound on $B$ in eq.~\eqref{eq:B-upper-bound-lem}.
	\subsection{Upper bounding $C$}
    \label{sec:upper-bounding-c}
	%
	%
	We express $C$ as below
	\begin{align}
		C & = \sum_{t=1}^T \left[\frac{1}{2}\norm{\bm{\Phi}(\bm{x}_t - \bm{z}_t^\star)}^2 + (\bm{y}_t - \bm{\Phi z}_t^\star)^\top \bm{\Phi}(\bm{z}_t^\star - \bm{x}_t)\right]\nonumber\\
		\ & = \sum_{t=1}^T \left[\frac{1}{2}\norm{\bm{\Phi}(\bm{x}_t - \bm{z}_t^\star)}^2 + \bm{w}_t^\top \bm{\Phi}(\bm{z}_t^\star - \bm{x}_t)\right.\nonumber\\
		\ & \left. + (\bm{u}_t - \bm{z}_t^\star)^\top \bm{\Phi}^\top\bm{\Phi}(\bm{z}_t^\star - \bm{x}_t) \right],
	\end{align}
	where we have used $\bm{y}_t = \bm{\Phi u}_t + \bm{w}_t$ in the last step. Using Cauchy-Schwartz inequality, we can further upper bound $C$ as below:
	\begin{align}
		C & \le \frac{P}{2} + Q + \sqrt{PS},
	\end{align}
	where 
        \begin{align}
	  \begin{aligned}
		\ & P = \sum_{t=1}^T \norm{\bm{\Phi}(\bm{x}_t - \bm{z}_t^\star)}^2, Q = \sum_{t=1}^T\bm{w}_t^\top \bm{\Phi}(\bm{z}_t^\star - \bm{x}_t),\\
		\ & S = \sum_{t=1}^T \norm{\bm{\Phi}(\bm{u}_t - \bm{z}_t^\star)}^2.
	\end{aligned}  
	\end{align}
    \paragraph{Upper bounding $Q$}
	We first derive a high probability upper bound on $Q$ in terms of $P$. In order to do this, we invoke the following Martingale Sub-Gaussian inequality~\cite[Theorem 13.3]{zhang2023mathematical}
	\begin{theorem}[Theorem 13.3,\cite{zhang2023mathematical}]
		\label{thm:martingale-subgaussian}
		Consider a sequence of random variables $\{Z_t\}$ and $S_t = [Z_1,\cdots, Z_t]$. Consider a sequence of random functions $\{\xi_i({S}_i)\}_i$. Assume that each $\xi_i$ is Sub-Gaussian with respect to $Z_i\vert S_{t-1}$, i.e.,
		\begin{align}
			\ln \mathbb{E}_{Z_i}[e^{\lambda\xi_i}\vert S_{t-1}] & \le \lambda \mathbb{E}_{Z_i}[\xi_i\vert S_{t-1}] + \frac{\lambda^2\sigma_i^2}{2},
		\end{align}
		for some $\sigma_i$ which may depend on $S_{i-1}.$ Then for all $\sigma>0$, with probability $\ge 1-\delta$, 
		\begin{align}
			\forall t>0: \sum_{i=1}^t\xi_i & \le \sum_{i=1}^t\mathbb{E}_{Z_i}[\xi_i\vert S_{t-1}]\nonumber\\
			\ & + \left(\sigma+\frac{\sum_{i=1}^t\sigma_i^2}{\sigma}\right)\sqrt{\frac{\ln(1/\delta)}{2}}. 
		\end{align}
	\end{theorem}
	To apply Theorem~\ref{thm:martingale-subgaussian}, we first observe that $\sum_{s=1}^t\bm{w}_s^\top \bm{\Phi}(\bm{z}_s^\star-\bm{x}_s)$ is a Martingale sequence with respect to the sequence $\{\bm{w}_t\}_{t\ge 1}$ since $\bm{x}_t$ is a function of $\bm{w}_1,\cdots, \bm{w}_{t-1}$, which is an i.i.d. sequence. Furthermore, observe that $\bm{w}_t^\top \bm{\Phi}(\bm{z}_t^\star-\bm{x}_t)$ is Sub-Gaussian with respect to $\bm{w}_1,\cdots, \bm{w}_{t-1}$ with Sub-Gaussian constant $\sigma_t = \norm{\bm{\Phi}(\bm{z}_t^\star-\bm{x}_t)}. $ Consequently, applying Theorem~\ref{thm:martingale-subgaussian}, we obtain, for any $\sigma>0$, w.p. $\ge 1-\delta$,
	\begin{align}
		\lefteqn{Q=\sum_{t=1}^T \bm{w}_t^\top \bm{\Phi}(\bm{z}_t^\star-\bm{x}_t)} & & \nonumber\\
		\ & \le \left(\sigma+\frac{\sum_{t=1}^T\norm{\bm{\Phi}(\bm{z}_t^\star-\bm{x}_t)}^2 }{\sigma}\right)\sqrt{\frac{\ln(1/\delta)}{2}}\nonumber\\
		\ & = \left(\sigma+\frac{P}{\sigma}\right)\sqrt{\frac{\ln(1/\delta)}{2}}.
	\end{align}
	As a result, we obtain, for any $\sigma>0$,
	\begin{align}
		C & \le \left(\sigma+\frac{P}{\sigma}\right)\sqrt{\frac{\ln(1/\delta)}{2}}+\frac{P}{2}+\sqrt{PS},\quad \mathrm{w.p.}\ge 1-\delta.
	\end{align}
	We can further bound $P,S$ using RIP as below:
	\begin{align}
		P & \le (1+\delta_K)P', & S& \le (1+\delta_K)S',
	\end{align}
	so that \begin{align}
    \label{eq:C-upper-bound-temp}
		C & \le \left(\sigma + \frac{(1+\delta_K)P'}{\sigma}\right)\sqrt{\frac{\ln(1/\delta)}{2}}\nonumber\\
		\ & + (1+\delta_K)\left(\frac{P'}{2}+\sqrt{P'S'}\right),\quad \mathrm{w.p.}\ge 1-\delta.
	\end{align}
	where \begin{align}
		P' & = \sum_{t=1}^T \norm{\bm{z}_t^\star-\bm{x}_t}^2, &
		S' & = \sum_{t=1}^T \norm{\bm{z}_t^\star-\bm{u}_t}^2.
	\end{align}
	Since $S'$ is deterministic, it is enough to obtain a high probability upper bound on $P'$. Since $P'$ depends on how $\bm{x}_t$ is updated, we expect different bounds on $P'$ for the agile and the lazy versions. We present a unified analysis below which considers the updates at a specific set of time instances, from which the result for both the agile and lazy versions will follow.
    \paragraph{A Generalized Analysis for Upper bounding $P'$}
    We consider the following generalization of the \texttt{FTASL}, where the update $\bm{x}_t$ is obtained only at a specific set of given time instances, denoted by $\mathcal{T}=\{t_k:0\le k\le k'-1\}$ such that $t_0=1$ and $t_{k'-1}\le T\le t_{k'}-1.$ Denoting $s_k = t_{k+1}-t_k,\ k\ge 0$, and using the inequality $\norm{\bm{a}+\bm{b}}^2\le 2(\norm{\bm{a}}+\norm{\bm{b}}^2)$, we obtain,
    \begin{align}
        \lefteqn{P' \le \sum_{k=0}^{k'-1}\sum_{t=t_k}^{t_{k+1}-1}\norm{\bm{z}_t^\star-\bm{x}_{t_k}}^2} & &\nonumber\\
        \ & \le 2\sum_{k=0}^{k'-1}s_k\norm{\bm{z}_{t_k - 1}^\star-\bm{x}_{t_k}}^2+2\sum_{k=0}^{k'-1}\sum_{t=t_k}^{t_{k+1}-1}\norm{\bm{z}_t^\star-\bm{z}_{t_k-1}^\star}^2\nonumber\\
        \ & =P_1''+P_2''.
    \end{align}
    Note that $P_1''$ is random but $P_2''$ is deterministic. We now obtain a high probability upper bound on $P_1''$. Observe that, for any $k\ge 0$, $\bm{x}_{t_k}$ is obtained by running \texttt{ALG} for $\tau_{t_k-1}$ iterations, with input measurement $\bm{b}_{t_k-1}$. Since $\bm{b}_{0}=\bm{0}$, by the algorithmic stability assumption, it follows that $\bm{x}_1 = \bm{0}$. Also, by definition, $\bm{z}^\star_0 = \bm{0}.$ Hence, $\norm{\bm{z}_0^\star - \bm{x}_1}=0$. Consequently, 
    \begin{align}
        P_1'' & = 2\sum_{k=1}^{k'-1}s_k\norm{\bm{z}_{t_k - 1}^\star-\bm{x}_{t_k}}^2.
    \end{align}
    For $k>0$, note that,
    \begin{align}
        \bm{b}_{t_k-1} & = \frac{\bm{Y}_{t_k-1}}{t_k-1}=\frac{\sum_{s=1}^{t_{k}-1}(\bm{\Phi u}_{s}+\bm{w}_s)}{t_k-1}\nonumber\\
        \ & = \bm{\Phi z}_{t_k-1}^\star + \bm{\nu}_{t_k-1}.
    \end{align}  Therefore, using the algorithmic stability assumption, we obtain, for $k>0$,
	\begin{align}
		\norm{\bm{x}_{t_k} - \bm{z}^\star_{t_k-1}} & \le 2^{-\tau_{t_k-1}}\norm{\bm{z}_{t_k-1}^\star} + \kappa\norm{\bm{\nu}_{t_k-1}}. 
	\end{align}
    Now, $\tau_{t_k-1}=\lceil\lg(t_k)\rceil=k$. Therefore,	for $k>0$,
	\begin{align}
		\norm{\bm{x}_{t_k} - \bm{z}^\star_{t_k-1}} & \le 2^{-k} + \kappa\norm{\bm{\nu}_{t_k-1}}. 
	\end{align}
	where we have used that for all $t\ge 1$, $\norm{\bm{z}_{t}^\star}=\frac{\norm{\sum_{s=1}^t\bm{u}_s}}{t}\le \frac{\sum_{s=1}^t \norm{\bm{u}_s}}{t}\le 1,$ since $\norm{\bm{u}_t}\le 1,\ \forall t\ge 1$ by the realizability assumption. This implies, 
    \begin{align}
        \lefteqn{P_1'' \le 2\sum_{k=1}^{k'-1}s_k(2^{-k} + \kappa\norm{\bm{\nu}_{t_k-1}})^2} & &\nonumber\\
        \ & \le 4\sum_{k=1}^{k'-1}s_k(2^{-2k}+\kappa^2\norm{\bm{\nu}_{t_k-1}}^2)\nonumber\\
        \ & \le 4\sum_{k=1}^{k'-1}s_k2^{-2k} + 4\kappa^2\sum_{k=1}^{k'-1}s_k\norm{\bm{\nu}_{t_k-1}}^2.
    \end{align}
    We now proceed to obtain a high probability upper bound on $\sum_{k=1}^{k'-1}s_k\norm{\bm{\nu}_{t_k-1}}^2$. Denote $\bm{e}_k = \bm{\nu}_{t_k-1},\ k=1,\cdots, k'-1$, so that we need a high probability upper bound on $\sum_{k=1}^{k'-1}s_k\norm{\bm{e}_k}^2$. 
    
    Denote, $t_{k'-1}-1=T_1$ and $\bm{x} = \begin{bmatrix}
		\bm{w}_1^\top & \bm{w}_2^\top & \cdots & \bm{w}_{T_1}^\top
	\end{bmatrix}^\top\in \real^{MT_1}$ and \begin{align}
		\bm{v}_k = 
		\begin{bmatrix}
			\underbrace{\begin{matrix}
					1 & \cdots & 1
			\end{matrix}}_{t_k-1} & \underbrace{\begin{matrix}
					0 & \cdots & 0
			\end{matrix}}_{T_1-t_k + 1}
		\end{bmatrix}^\top\in \real^{T_1},\ 1\le k\le k'-1.
	\end{align}  Note that \begin{align}
	    \sum_{k=1}^{k'-1}s_k\norm{\bm{e}_k}^2 & = \sum_{k=1}^{k'-1}s_k\norm{\frac{\sum_{t=1}^{t_k-1}\bm{w}_t}{t_{k}-1}}^2\nonumber\\
        \ & = \sum_{k=1}^{k'-1}\frac{s_k}{(t_k-1)^2}\bm{x}^\top (\bm{v}_k\otimes \bm{I}_M)(\bm{v}_k\otimes \bm{I}_M)^\top\bm{x}\nonumber\\
        \ & =\bm{x}^\top \bm{A}\bm{x},
    \end{align} 
    where \begin{align}
        \bm{A} & =\sum_{k=1}^{k'-1}\bm{a}_k\bm{a}_k^\top, & \bm{a}_k & = \frac{\sqrt{s_k}\bm{v}_k\otimes \bm{I}_M}{t_k-1},\ k=1,\cdots, k'-1.
    \end{align}
    To further proceed, we invoke the following result, whose proof can be found in the extended technical report:
    \begin{lem}
        \label{lem:psd-hanson-wright}
        Let $\bm{x}\in \real^n$ be such that $\bm{x}\sim\mathcal{N}(\bm{0},\bm{I}_n)$. Then, for any matrix $\bm{A}=\sum_{l=1}^L \bm{a}_l\bm{a}_l^\top$, the following concentration inequality holds:
        \begin{align}
            \label{eq:psd-hanson-wright}
		\prob{\bm{x}^\top \bm{Ax} \le b(\delta)\sum_{l=1}^L\norm{\bm{a}_l}^2} & \ge 1-\delta,
        \end{align}
        where $b(\delta) = M+2\sqrt{M\ln(1/\delta)}+2\ln(1/\delta)=(\sqrt{M}+\sqrt{\ln(1/\delta)})^2+\ln(1/\delta)$. 
    \end{lem}
    \begin{proof}
        Since $\bm{A}$ is PSD, it admits an eigenvalue decomposition $\bm{A}=\bm{VDV}^\top$, where $\bm{D}$ is diagonal with the diagonal entries consisting of eigenvalues of $\bm{A}$ and $\bm{V}$ is the matrix with columns as the normalized eigenvectors of $\bm{A}$. Then, $\bm{x}^\top \bm{Ax} = \bm{y}^\top \bm{D y},$
	where $\bm{y}=\bm{Vx}$. Since $\bm{V}$ is unitary, $\bm{y}\sim \mathcal{N}(\bm{0},\bm{I}_{n})$. Let $\bm{d}$ be the vector of eigenvalues of $\bm{A}$. Then, $\bm{y}^\top \bm{D y}=\sum_{i=1}^{n}d_iy_i^2$. By Lemma 1 of~\cite{laurent2000adaptive}, we can obtain the following concentration of $\sum_{i=1}^{n}d_iy_i^2$, for any $\delta\in (0,1)$:
	\begin{align}
		\prob{\sum_{i=1}^{n}d_iy_i^2 \le a(\delta)} & \ge 1-\delta,
	\end{align}
	where $a(\delta)=\opnorm{\bm{d}}{1}+2\norm{\bm{d}}\sqrt{\ln(1/\delta)}+2\opnorm{\bm{d}}{\infty}\ln(1/\delta)$.
	It suffices to obtain upper bounds of $\opnorm{\bm{d}}{1},\norm{\bm{d}}$ and $\opnorm{\bm{d}}{\infty}.$ Since $\bm{A}$ is PSD, the eigenvalues are non-neagtive, so that We obtain, 
	\begin{align}
		\opnorm{\bm{d}}{1} & =\mathrm{Tr}(\bm{A})=\sum_{l=1}^{L} \mathrm{Tr}\left(\bm{a}_l\bm{a}_l^\top\right) = \sum_{l=1}^{L}\norm{\bm{a}_l}^2. 
	\end{align}
	Furthermore, 
	\begin{align}
		\norm{\bm{d}} & =\opnorm{\bm{A}}{\mathrm{F}} \le \sum_{l=1}^{L}\opnorm{\bm{a}_l\bm{a}_l^\top}{F}=\sum_{l=1}^{L}\norm{\bm{a}_l}^2. 
	\end{align}
	Moreover,
	\begin{align}
		\opnorm{\bm{d}}{\infty} & = \opnorm{\bm{A}}{2\to 2}\le \sum_{l=1}^{L}\opnorm{\bm{a}_l\bm{a}_l^\top}{2\to2}=\sum_{l=1}^{L}\norm{\bm{a}_l}^2.
	\end{align}
	Consequently, we obtain, 
	\begin{align}
		\prob{\bm{x}^\top \bm{Ax} \le b(\delta)\sum_{l=1}^L\norm{\bm{a}_l}^2} & \ge 1-\delta,
	\end{align}
	where $b(\delta) = M+2\sqrt{M\ln(1/\delta)}+2\ln(1/\delta)=(\sqrt{M}+\sqrt{\ln(1/\delta)})^2+\ln(1/\delta)$.
    \end{proof}
   Using the Lemma~\ref{lem:psd-hanson-wright}, we obtain that, w.p. $\ge 1-\delta$, $\sum_{k=1}^{k'-1}s_k\norm{\bm{e}_k}^2\le b(\delta)\sum_{k=1}^{k'-1}\norm{\bm{a}_k}^2$. Now, 
   \begin{align}
   \label{eq:a-norm-temp}
       \sum_{k=1}^{k'-1}\norm{\bm{a}_k}^2 & = \sum_{k=1}^{k'-1}\frac{Ms_k\norm{\bm{v}_k}^2}{(t_k-1)^2}=M\sum_{k=1}^{k'-1}\frac{s_k}{t_k-1}.
   \end{align}
   For \texttt{L-FTASL}, $t_k=2^k,k=0,\cdots, k'-1$, where $k'=\lceil\lg(T+1)\rceil.$ Therefore, $s_k = t_{k+1}-t_k=2^{k+1}-2^k=2^k,\ k\ge 1$. Therefore, we obtain, from Eq.~\eqref{eq:a-norm-temp} that
   \begin{align}
   \label{eq:a-norm-temp-l-ftasl}
       \sum_{k=1}^{k'-1}\norm{\bm{a}_k}^2 & = M\sum_{k=1}^{k'-1}\frac{2^k}{2^k-1}\le 2Mk'=\mathcal{O}(M\ln T).
   \end{align}

   For \texttt{A-FTASL}, we have $t_k=k,\ k=0,\cdots, T$, so that $s_k=1,\forall k$. Then, from Eq.~\eqref{eq:a-norm-temp} we obtain,
   \begin{align}
   \label{eq:a-norm-temp-a-ftasl}
       \sum_{k=1}^{k'-1}\norm{\bm{a}_k}^2 & = M\sum_{t=2}^{T}\frac{1}{t-1}\le M\ln(eT)=\mathcal{O}(M\ln T).
   \end{align}
   Furthermore, for \texttt{L-FTASL},
   \begin{align}
       \sum_{k=1}^{k'-1}s_k2^{-2k}=\sum_{k=1}^{k'-1}2^{-k}\le 1.
   \end{align}
   And, for \texttt{A-FTASL}, 
   \begin{align}
   \sum_{k=1}^{k'-1}s_k2^{-2k}=\sum_{t=1}^{T}2^{-2t}\le \frac{1}{3}.
   \end{align}
   Hence, we obtain, for both \texttt{A-FTASL} and \texttt{L-FTASL} that,
   $P_1'' = 1+2\kappa^2 Mb(\delta)\ln T$, w.p.$\ge 1-\delta$. 

   Hence, we obtain that 
   \begin{align}
       P' & \le 1+2\kappa^2 Mb(\delta)\ln T \nonumber\\
                 \label{eq:P'-upper-bound}
       \ &  + 
2\sum_{k=0}^{k'-1}\sum_{t=t_k}^{t_{k+1}-1}\norm{\bm{z}_t^\star-\bm{z}_{t_k-1}^\star}^2=:a_T(\delta).
   \end{align} 
	Using this, we obtain from~\eqref{eq:C-upper-bound-temp} that, for any $\sigma>0$, w.p. $\ge 1-\delta$,
	\begin{align}
		C & \le \left(\sigma + \frac{(1+\delta_K)a_T(\delta)}{\sigma}\right)\sqrt{\frac{\ln(2/\delta)}{2}}\nonumber\\
		\ & + (1+\delta_K)\left(\frac{a_T(\delta)}{2}+\sqrt{a_T(\delta)S'}\right).
	\end{align}
	Choosing $\sigma=\sqrt{(1+\delta_{K})a_T(\delta)}$, we obtain w.p.$\ge 1-\delta$, 
	\begin{align}
		\label{eq:C-bound-a-ftasl}
		C & \le \sqrt{2\ln(2/\delta)(1+\delta_K)a_T(\delta)}\nonumber\\
        \ & + (1+\delta_K)\left(\frac{a_T(\delta)}{2}+\sqrt{a_T(\delta)S'}\right).
	\end{align}
\bibliographystyle{IEEEtran}
\bibliography{osa_bib}

\begin{thebibliography}{10}
\providecommand{\url}[1]{#1}
\csname url@samestyle\endcsname
\providecommand{\newblock}{\relax}
\providecommand{\bibinfo}[2]{#2}
\providecommand{\BIBentrySTDinterwordspacing}{\spaceskip=0pt\relax}
\providecommand{\BIBentryALTinterwordstretchfactor}{4}
\providecommand{\BIBentryALTinterwordspacing}{\spaceskip=\fontdimen2\font plus
\BIBentryALTinterwordstretchfactor\fontdimen3\font minus
  \fontdimen4\font\relax}
\providecommand{\BIBforeignlanguage}[2]{{%
\expandafter\ifx\csname l@#1\endcsname\relax
\typeout{** WARNING: IEEEtran.bst: No hyphenation pattern has been}%
\typeout{** loaded for the language `#1'. Using the pattern for}%
\typeout{** the default language instead.}%
\else
\language=\csname l@#1\endcsname
\fi
#2}}
\providecommand{\BIBdecl}{\relax}
\BIBdecl

\bibitem{tibshirani1996regression}
R.~Tibshirani, ``Regression shrinkage and selection via the lasso,''
  \emph{Journal of the Royal Statistical Society Series B: Statistical
  Methodology}, vol.~58, no.~1, pp. 267--288, 1996.

\bibitem{chen2001atomic}
S.~S. Chen, D.~L. Donoho, and M.~A. Saunders, ``Atomic decomposition by basis
  pursuit,'' \emph{SIAM review}, vol.~43, no.~1, pp. 129--159, 2001.

\bibitem{candes2007dantzig}
E.~Candes and T.~Tao, ``The dantzig selector: Stastical estimation when $p$ is
  much larger than $n$.'' \emph{Annals of statistics}, vol.~35, no.~6, pp.
  2313--2351, 2007.

\bibitem{daubechies2004iterative}
I.~Daubechies, M.~Defrise, and C.~De~Mol, ``An iterative thresholding algorithm
  for linear inverse problems with a sparsity constraint,''
  \emph{Communications on Pure and Applied Mathematics: A Journal Issued by the
  Courant Institute of Mathematical Sciences}, vol.~57, no.~11, pp. 1413--1457,
  2004.

\bibitem{dai2009subspace}
W.~Dai and O.~Milenkovic, ``Subspace pursuit for compressive sensing signal
  reconstruction,'' \emph{IEEE transactions on Information Theory}, vol.~55,
  no.~5, pp. 2230--2249, 2009.

\bibitem{needell2009cosamp}
D.~Needell and J.~A. Tropp, ``Cosamp: Iterative signal recovery from incomplete
  and inaccurate samples,'' \emph{Applied and computational harmonic analysis},
  vol.~26, no.~3, pp. 301--321, 2009.

\bibitem{foucart2011hard}
S.~Foucart, ``Hard thresholding pursuit: an algorithm for compressive
  sensing,'' \emph{SIAM Journal on numerical analysis}, vol.~49, no.~6, pp.
  2543--2563, 2011.

\bibitem{blanchard2015performance}
J.~D. Blanchard and J.~Tanner, ``Performance comparisons of greedy algorithms
  in compressed sensing,'' \emph{Numerical Linear Algebra with Applications},
  vol.~22, no.~2, pp. 254--282, 2015.

\bibitem{kale2017adaptive}
S.~Kale, Z.~Karnin, T.~Liang, and D.~P{\'a}l, ``Adaptive feature selection:
  Computationally efficient online sparse linear regression under rip,'' in
  \emph{International Conference on Machine Learning}.\hskip 1em plus 0.5em
  minus 0.4em\relax PMLR, 2017, pp. 1780--1788.

\bibitem{bhattacharjee2020fundamental}
R.~Bhattacharjee, S.~Banerjee, and A.~Sinha, ``Fundamental limits on the regret
  of online network-caching,'' \emph{Proceedings of the ACM on Measurement and
  Analysis of Computing Systems}, vol.~4, no.~2, pp. 1--31, 2020.

\bibitem{mukhopadhyay2021online}
S.~Mukhopadhyay and A.~Sinha, ``Online caching with optimal switching regret,''
  in \emph{2021 IEEE International Symposium on Information Theory
  (ISIT)}.\hskip 1em plus 0.5em minus 0.4em\relax IEEE, 2021, pp. 1546--1551.

\bibitem{chen2017}
T.~Chen, Q.~Ling, and G.~B. Giannakis, ``An online convex optimization approach
  to proactive network resource allocation,'' \emph{IEEE Trans. Signal
  Process.}, vol.~65, no.~24, pp. 6350--6364, 2017.

\bibitem{ziniel2013dynamic}
J.~Ziniel and P.~Schniter, ``Dynamic compressive sensing of time-varying
  signals via approximate message passing,'' \emph{IEEE Trans. Signal
  Process.}, vol.~61, no.~21, pp. 5270--5284, 2013.

\bibitem{asif2014sparse}
M.~S. Asif and J.~Romberg, ``Sparse recovery of streaming signals using
  $l_1$-homotopy,'' \emph{IEEE Trans. Signal Process.}, vol.~62, no.~16, pp.
  4209--4223, 2014.

\bibitem{vaswani2016recursive}
N.~Vaswani and J.~Zhan, ``Recursive recovery of sparse signal sequences from
  compressive measurements: A review,'' \emph{IEEE Trans. Signal Process.},
  vol.~64, no.~13, pp. 3523--3549, 2016.

\bibitem{fosson2017online}
S.~M. Fosson, J.~Matamoros, M.~Gregori, and E.~Magli, ``Online convex
  optimization meets sparsity,'' in \emph{SPARS workshop}, 2017.

\bibitem{fosson2020centralized}
S.~M. Fosson, ``Centralized and distributed online learning for sparse
  time-varying optimization,'' \emph{IEEE Transactions on Automatic Control},
  vol.~66, no.~6, pp. 2542--2557, 2020.

\bibitem{foster2016online}
D.~Foster, S.~Kale, and H.~Karloff, ``Online sparse linear regression,'' in
  \emph{Conference on Learning Theory}.\hskip 1em plus 0.5em minus 0.4em\relax
  PMLR, 2016, pp. 960--970.

\bibitem{ito2017efficient}
S.~Ito, D.~Hatano, H.~Sumita, A.~Yabe, T.~Fukunaga, N.~Kakimura, and K.-I.
  Kawarabayashi, ``Efficient sublinear-regret algorithms for online sparse
  linear regression with limited observation,'' \emph{Advances in Neural
  Information Processing Systems}, vol.~30, 2017.

\bibitem{ito2018online}
------, ``Online regression with partial information: Generalization and linear
  projection,'' in \emph{International Conference on Artificial Intelligence
  and Statistics}.\hskip 1em plus 0.5em minus 0.4em\relax PMLR, 2018, pp.
  1599--1607.

\bibitem{wang2019online}
J.-K. Wang, C.-J. Lu, and S.-D. Lin, ``Online linear optimization with sparsity
  constraints,'' in \emph{Algorithmic Learning Theory}.\hskip 1em plus 0.5em
  minus 0.4em\relax PMLR, 2019, pp. 883--897.

\bibitem{mukhopadhyay2022k}
S.~Mukhopadhyay, S.~Sahoo, and A.~Sinha, ``k-experts-online policies and
  fundamental limits,'' in \emph{International Conference on Artificial
  Intelligence and Statistics}.\hskip 1em plus 0.5em minus 0.4em\relax PMLR,
  2022, pp. 342--365.

\bibitem{kalai2005efficient}
A.~Kalai and S.~Vempala, ``Efficient algorithms for online decision problems,''
  \emph{Journal of Computer and System Sciences}, vol.~71, no.~3, pp. 291--307,
  2005.

\bibitem{davis1997adaptive}
G.~Davis, S.~Mallat, and M.~Avellaneda, ``Adaptive greedy approximations,''
  \emph{Constructive approximation}, vol.~13, pp. 57--98, 1997.

\bibitem{zhao2023improved}
Y.~Zhao and Z.~Luo, ``Improved rip-based bounds for guaranteed performance of
  two compressed sensing algorithms,'' \emph{Science China Mathematics},
  vol.~66, no.~5, pp. 1123--1140, 2023.

\bibitem{blumensath08iht}
T.~Blumensath and M.~E. Davies, ``Iterative hard thresholding for compressed
  sensing,'' \emph{Applied and Computational Harmonic Analysis}, vol.~27,
  no.~3, pp. 265--274, 2009.

\bibitem{foucart2013}
S.~Foucart and H.~Rauhut, \emph{A Mathematical Introduction to Compressive
  Sensing}.\hskip 1em plus 0.5em minus 0.4em\relax New York, NY: Springer New
  York, 2013.

\bibitem{boucheron2003concentration}
S.~Boucheron, G.~Lugosi, and O.~Bousquet, ``Concentration inequalities,'' in
  \emph{Summer school on machine learning}.\hskip 1em plus 0.5em minus
  0.4em\relax Springer, 2003, pp. 208--240.

\bibitem{alpaydin1998pen}
E.~Alpaydin and F.~Alimoglu, ``Pen-based recognition of handwritten digits data
  set. university of california, irvine,'' \emph{Machine Learning Repository.
  Irvine: University of California}, vol.~4, no.~2, 1998.

\bibitem{laurent2000adaptive}
B.~Laurent and P.~Massart, ``Adaptive estimation of a quadratic functional by
  model selection,'' \emph{Annals of statistics}, pp. 1302--1338, 2000.

\bibitem{zhang2023mathematical}
T.~Zhang, \emph{Mathematical analysis of machine learning algorithms}.\hskip
  1em plus 0.5em minus 0.4em\relax Cambridge University Press, 2023.

\end{thebibliography}
\end{document}